\documentclass[11pt]{article}

\usepackage[preprint]{acl}

\usepackage{times}
\usepackage{latexsym}
\usepackage{enumitem}

\usepackage[T1]{fontenc}
\usepackage[utf8]{inputenc}

\usepackage{microtype}

\usepackage{inconsolata}

\usepackage{graphicx}

\usepackage{amsmath}
\usepackage{amssymb}
\usepackage{booktabs}
\usepackage{wrapfig}
\usepackage{xcolor}
\usepackage[most]{tcolorbox}
\usepackage{CJKutf8}
\usepackage[table]{xcolor}
\definecolor{cueblue}{RGB}{235,244,252}

\newtcolorbox{promptbox}[1][]{
    breakable,
    colback=gray!4,
    colframe=gray!45,
    boxrule=0.5pt,
    arc=1pt,
    left=4pt,
    right=4pt,
    top=4pt,
    bottom=4pt,
    fontupper=\footnotesize\ttfamily,
    fonttitle=\small\bfseries,
    title=#1
}

\title{Retrieval-Enhanced Cultural Evaluation for Social Media Translation Training}

\author{
  \textbf{Linjuan Wu\textsuperscript{1,*}},
  \textbf{Yiwen Qiu\textsuperscript{1,2,*}},
  \textbf{Xinze Lyu\textsuperscript{1}},
  \textbf{Dingming Li\textsuperscript{1,2}},
  \textbf{Hui Yao\textsuperscript{1}},
  \\
  \textbf{Zhongyi Sang\textsuperscript{1}},
  \textbf{Ye Guo\textsuperscript{1}},
  \textbf{Daoxin Zhang\textsuperscript{1}},
  \textbf{Zhe Xu\textsuperscript{1}},
  \textbf{Fei Zhao\textsuperscript{1}},
  \\
  \textbf{Mu Chuan\textsuperscript{1}},
  \textbf{Yao Hu\textsuperscript{1}},
  \textbf{Weiming Lu\textsuperscript{2,$\dagger$}},
  \textbf{Yongliang Shen\textsuperscript{2,$\dagger$}}
  \\
  \\
  \textsuperscript{1}Xiaohongshu Inc.
  \quad
  \textsuperscript{2}Zhejiang University
  \\
  \texttt{wulinjuan@xiaohongshu.com}
}

\begin{document}
\maketitle

% Place immediately after \maketitle
% \footnotetext[*]{Equal contribution.}
% \footnotetext[$\dagger$]{Corresponding authors.}

\begingroup
\renewcommand{\thefootnote}{\fnsymbol{footnote}}
\footnotetext[1]{Equal contribution.}
\footnotetext[2]{Corresponding authors.}
\endgroup

\begin{abstract}
Social media translation requires preserving meaning while capturing the cultural nuances of user-generated content (UGC). However, gaps in the cultural knowledge of LLM-based evaluators can lead to inaccurate assessments and unreliable training rewards. To improve these signals, we use retrieval-enhanced cultural evaluation to train \textbf{CUE-MT}, a family of social media translation models. A calibrated, domain-specific Judger guides supervised data refinement and is augmented by an Agent for hard-example mining and reinforcement learning. The Agent retrieves relevant external evidence as needed and distills it into concise hints that the Judger uses to score translations and assign rewards. Across 9B, 27B, and 35B-A3B backbones, CUE-MT outperforms the evaluated open-source baselines of comparable size on CUE-MT-Bench and three public benchmarks. Human evaluation on CUE-MT-Bench shows cultural effectiveness gains of 7.44–15.12 percentage points over the strongest external baseline in each size group. Retrieval improves the Judger’s four-way classification accuracy by 8.07 percentage points over a matched no-retrieval baseline that uses LLM-generated hints alone. Training ablations further support the combination of hard-example mining and Agent-enhanced rewards.

\end{abstract}

\section{Introduction}
Social media is a major channel for cross-lingual communication and cultural exchange \citep{lin2018mining, ugc1, ugc2}. Compared with news and encyclopedic text \citep{yao2024benchmarking}, social media language is more informal and context-dependent, often featuring slang, memes, and community-specific jargon \citep{baldwin2013noisy, mei2024slang, wuraola2024slang, slangdit2025, huang2025buzzwords}. Translating such content requires more than literal accuracy; translations must also convey cultural meaning, emotional tone, and communicative intent \citep{guo2025redefiningmachinetranslationsocial, zhang2023sociopragmatic, qiu2026metricsrewardworsttranslations}.

CULTURE-MT \citep{CULTURE-MT} introduced cultural effectiveness as an evaluation criterion and trained a dedicated Judger that outperforms lexical metrics \citep{papineni2002bleu, popovic2015chrf} and LLM judges \citep{kocmi2023gemba}. CuRIL \citep{qiu2026metricsrewardworsttranslations} further showed that evaluation errors of Judger can stem from cultural knowledge gaps, not just reasoning limitations. It found that cultural hints supplied at evaluation time improve agreement with human judgments and proposed a training procedure to internalize cultural reasoning.  However, training may not fully cover the diverse and evolving cultural knowledge of social media~\citep{dhingra2022timeaware}. These knowledge gaps can lead to inaccurate evaluations and distort the rewards used to train translation models through reinforcement learning \citep{shen2016mrt, stiennon2020summarize, anon2026grrm}.

\begin{figure*}[h]
\centering
\includegraphics[width=\textwidth]{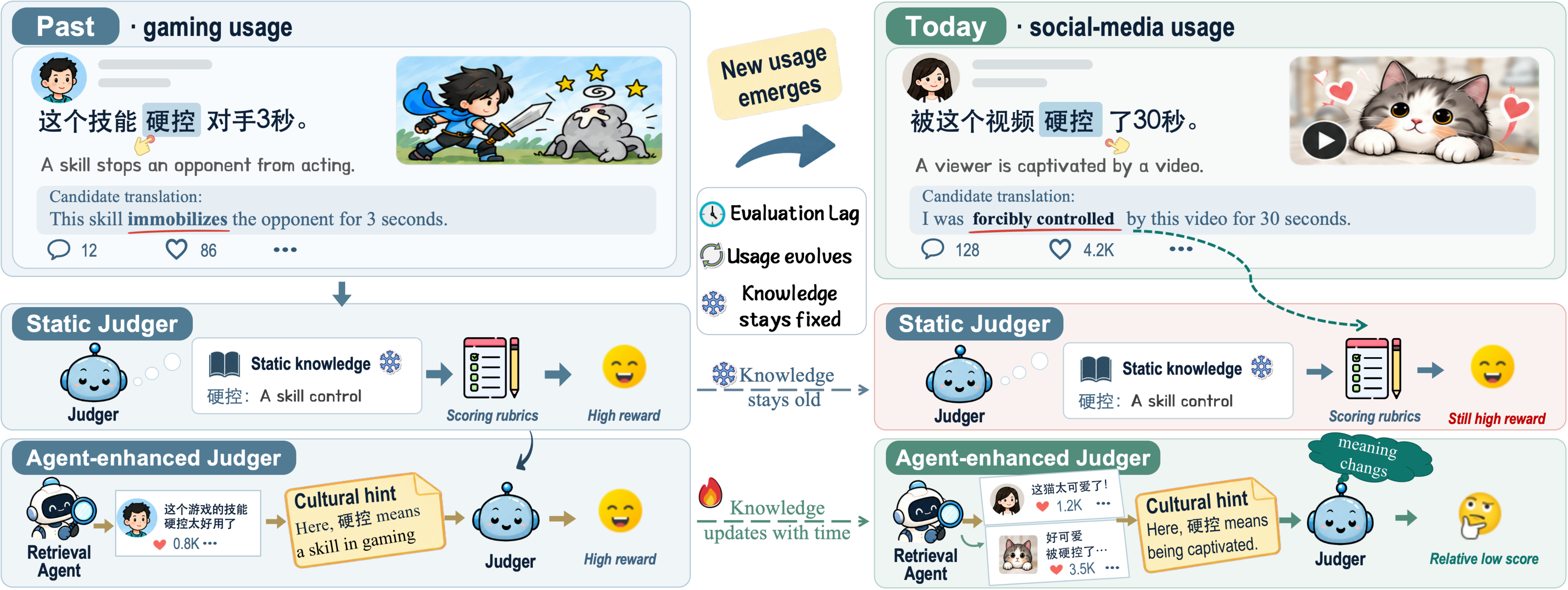}
\caption{
An example of a cultural knowledge gap in translation evaluation. ``\begin{CJK*}{UTF8}{gbsn}硬控\end{CJK*}'' refers to immobilizing an opponent in gaming, but it can also mean being captivated by content on social media. Applying the gaming meaning may cause a static Judger to overrate ``forcibly controlled''. Retrieved evidence helps the Agent-enhanced Judger interpret the expression in context and assign a more appropriate score.
}
\label{fig:motivation}
\end{figure*}

We address \textbf{cultural knowledge gaps in evaluation} with an \textbf{Agent-enhanced Judger}. These gaps arise when a Judger lacks the cultural knowledge needed to assess a translation or fails to apply it correctly in context (Figure~\ref{fig:motivation}). The Agent identifies expressions requiring cultural knowledge in each input, retrieves relevant external evidence as needed, and summarizes it into concise hints. A calibrated, domain-specific Judger uses these hints to assign the final scores.

We use this evaluation framework to train \textbf{CUE-MT}, a family of social media translation models at 9B, 27B, and 35B-A3B scales. The static Judger guides supervised data refinement, while the Agent-enhanced Judger scores sampled translations for hard-example mining and provides cultural-effectiveness rewards for reinforcement learning. Across all three scales, CUE-MT outperforms the evaluated open-source baselines of comparable size on CUE-MT-Bench, CULTURE-MT~\citep{CULTURE-MT}, Seed-X-Challenge~\citep{cheng2025seedx}, and RedTrans~\citep{guo2025redefiningmachinetranslationsocial}. Human evaluation on CUE-MT-Bench confirms gains in cultural effectiveness. The Agent-enhanced Judger also outperforms the evaluated static and LLM judges. Ablations show that retrieval improves Judger accuracy beyond generated hints alone and that Agent-enhanced rewards improve GRPO results over static rewards.

Our main contributions are as follows:
\begin{itemize}
    \item \textbf{Cultural-effectiveness-guided translation training.}
    We introduce \textbf{CUE-MT}, a family of social media translation models trained using cultural-effectiveness feedback for supervised data refinement, hard-example mining, and reinforcement learning.
    \item \textbf{Retrieval-enhanced cultural evaluation.}
    We propose an Agent-enhanced Judger in which an Agent retrieves external cultural evidence and provides concise hints to a calibrated, domain-specific Judger for translation scoring.
    \item \textbf{Cross-scale empirical validation.}
    CUE-MT achieves the best overall performance among the evaluated open-source models of comparable size at 9B, 27B, and 35B-A3B scales. Human evaluation confirms gains in cultural effectiveness, while controlled ablations support the benefits of retrieval and Agent-enhanced rewards.
\end{itemize}

\section{Related Work}

\subsection{Social Media and Culture-Aware Translation}

Research on social media translation addresses both noisy language and culturally specific expressions. MTNT \citep{michel2018mtnt} provides a benchmark of naturally noisy Reddit comments with professional translations. Subsequent work examines informal language, Internet slang, and buzzwords \citep{zhang2023sociopragmatic,mei2024slang,wuraola2024slang,huang2025buzzwords,sun2024informal}. Translation-focused studies address slang \citep{slangdit2025} and broader social media phenomena, including humor, emojis, and memes \citep{guo2025redefiningmachinetranslationsocial}. Culture-aware translation research further examines how cultural meaning transfers across languages. CAMT benchmarks the translation of culture-specific items, including those without direct target-language equivalents \citep{yao2024benchmarking}. KG-MT retrieves information from multilingual knowledge graphs to improve the translation of culturally nuanced entity names \citep{conia2024towards}. CULTURE-MT introduces a social media translation benchmark and the criterion of \emph{cultural effectiveness}, which considers both expression accuracy and cultural adaptability \citep{CULTURE-MT}.

\begin{figure*}[h]
    \centering
    \includegraphics[width=\textwidth]{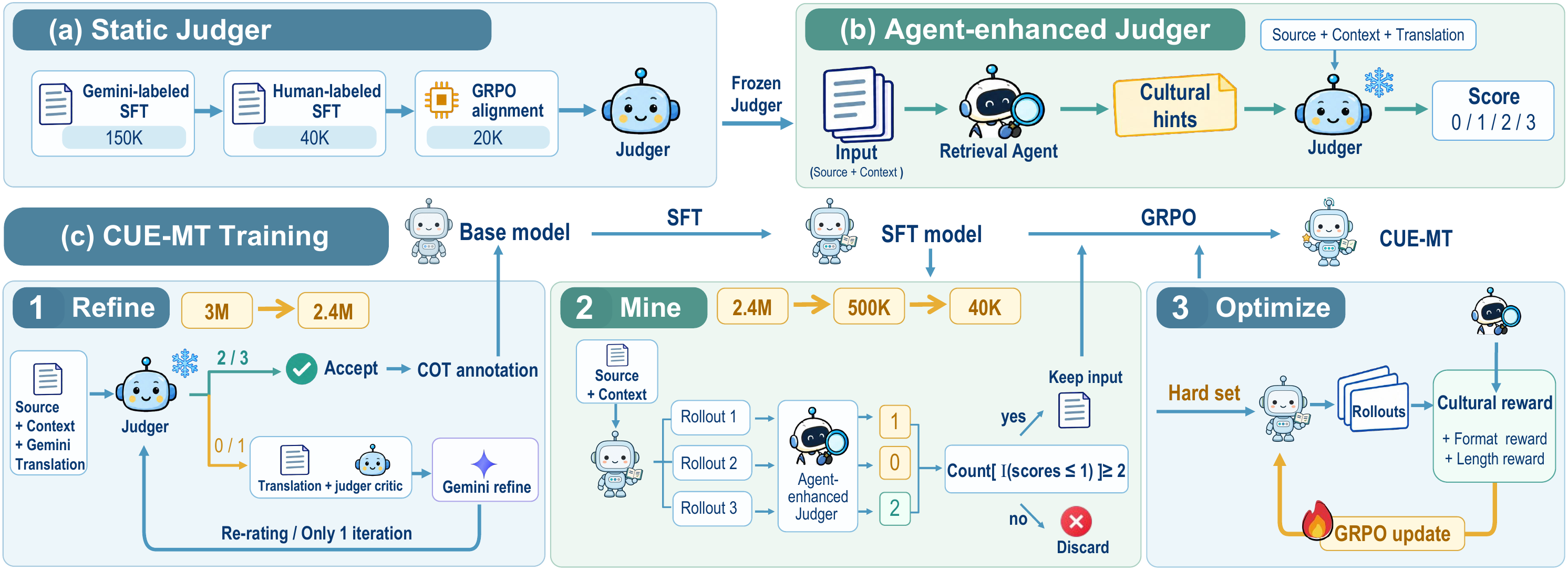}
    \caption{
    Overview of CUE-MT training. The static Judger guides supervised data refinement (Refine). The Agent-enhanced Judger uses external cultural evidence to score translations for hard-example mining (Mine) and provide cultural-effectiveness rewards during GRPO (Optimize).
    }
    \label{fig:framework}
\end{figure*}

\subsection{Translation Evaluation and External Knowledge}

Automatic MT evaluation includes lexical metrics such as BLEU \citep{papineni2002bleu} and chrF \citep{popovic2015chrf}, learned metrics such as BLEURT \citep{sellam2020bleurt}, COMET \citep{rei2020comet}, and xCOMET \citep{guerreiro2024xcomet}, and prompt-based LLM judges such as GEMBA \citep{kocmi2023gemba}. For cultural-effectiveness assessment, CULTURE-MT trains a dedicated Judger \citep{CULTURE-MT}. CuRIL identifies cultural knowledge gaps as a source of evaluation errors and uses hint-guided reinforcement learning to internalize cultural reasoning. It also applies the trained Judger as a reward model for downstream translation optimization \citep{qiu2026metricsrewardworsttranslations}. Retrieval-augmented generation provides access to external knowledge \citep{lewis2020rag}, while tool-using language models can learn when to invoke external resources \citep{schick2023toolformer}. More directly related to translation evaluation, RATE coordinates search, evaluation, and comparison agents to assess non-literal translations \citep{tian2026beyond}. Our work complements cultural-knowledge internalization with input-dependent retrieval for translation training. A calibrated, domain-specific Judger guides supervised data refinement, while an Agent supplies it with concise hints derived from retrieved evidence for hard-example mining and reinforcement-learning rewards.

\section{Method}
\label{sec:method}

We train CUE-MT using cultural-effectiveness feedback from two configurations of the same Judger: a static configuration for supervised data refinement and an Agent-enhanced configuration for hard-example mining and reinforcement learning (Figure~\ref{fig:framework}). 
We first describe how the Judger is trained and enhanced with retrieval (Section~\ref{sec:agent-judger}), and then present the translation training pipeline (Section~\ref{sec:training-pipeline}).

\subsection{Agent-enhanced Judger}
\label{sec:agent-judger}

\paragraph{Judger Training.}

We translate source texts from publicly available social media data using the baseline models described in Section~\ref{sec:baselines}, including both specialized translation models and general-purpose LLMs. This yields 300K source--translation pairs, of which 200K are annotated by Gemini and 100K by human annotators. For initial supervised fine-tuning, we select 150K Gemini-annotated pairs balanced across the four cultural-effectiveness scores $\{0,1,2,3\}$. We then perform second-stage SFT on 40K human-annotated pairs balanced across the same scores.

From this 40K human-annotated set, we sample 20K examples for Judger GRPO training, reserving 5\% for GRPO validation. Let $s^\star$ denote the human label, $\hat{s}$ the predicted score, and $b(s)=\mathbf{1}[s\geq 2]$ the corresponding binary acceptance label. The Judger reward combines binary and four-way correctness:
\begin{equation*}
r_{\mathrm{J}}
=
\mathbf{1}[b(\hat{s})=b(s^\star)]
+
0.5\,\mathbf{1}[\hat{s}=s^\star].
\end{equation*}
This rewards both culturally effective decisions and exact agreement with human scores. For example, an exact four-way match receives 1.5 points, whereas a correct binary class with an incorrect four-way score receives 1 point. The four-way component explicitly supervises distinctions within the coarse classes. Model configurations and training details are provided in Appendix~\ref{app:training-details-judger}.

\paragraph{Static Scoring.}

Given a source text $x$, optional context $c$, and candidate translation $y$, the static Judger predicts the score $\hat{s}$:
\begin{equation*}
\hat{s}=f_{\theta}(x,c,y),
\qquad
\hat{s}\in\{0,1,2,3\}.
\end{equation*}

\paragraph{Agent Enhancement.}
To supplement the knowledge available for scoring, an Agent plans input-specific queries, retrieves external evidence, and summarizes it into concise hints for the Judger:
\begin{align*}
Q &= g_{\mathrm{plan}}(x,c,y),\\
D &= \mathcal{R}(Q),\\
H &= g_{\mathrm{hint}}(x,c,y,Q,D),\\
\hat{s} &= f_{\theta}(x,c,y,H),
\end{align*}
where $Q$ denotes retrieval queries, $D$ is the retrieved evidence, and $H$ the judging hints.

If the planner generates no queries, the Judger scores the translation without hints. If retrieval yields no useful evidence or fails, the Agent instead provides general judging guidance without external evidence. Implementation details are provided in Appendix~\ref{app:retrieval-execution}. Appendix~\ref{app:agent-prompts} provides the planning, hint-generation, and hint-injection prompts.

\subsection{Social Media Translation Training}
\label{sec:training-pipeline}

CUE-MT training consists of supervised data construction and fine-tuning, hard-example mining, and GRPO~\citep{grpo}.

\paragraph{Supervised Data Construction.}

We sample 3 million Chinese social media instances from public platforms. This collection does not overlap with the Judger-training data. We use Gemini-3.1-Pro~\citep{google-gemini-3} to generate English translations. The static Judger evaluates all translations without invoking the Agent. Translations scored 2 or 3 are retained directly. Those scored 0 or 1 are revised once by Gemini-3.1-Pro using Judger feedback and retained only if the static Judger assigns a score of 2 or 3 upon re-evaluation. The prompt for refining is shown in Appendix~\ref{app:translation-refinement}. This process yields 2.4 million source--translation pairs.

\paragraph{Supervised Fine-Tuning.}

To provide explicit supervision for interpreting context and cultural expressions, we use GLM-5 to construct a concise chain-of-thought (CoT) for each retained pair using the prompt in Appendix~\ref{app:cot-generation}. The CoT comprises \emph{context analysis}, which summarizes relevant context, and \emph{translation points}, which explain how to convey key expressions and cultural nuances in English. We combine the generated CoT with the retained translation to form the SFT target. Given the source text and optional context, the model is fine-tuned to generate the CoT followed by the final English translation, using the translation prompt in Appendix~\ref{app:translation-prompts}.

\paragraph{Hard-Example Mining.}

To focus GRPO training on cultural and contextual challenges that remain after SFT, we mine hard examples separately for the 9B, 27B, and 35B-A3B models. We randomly sample 500K source instances from the original pool of 3 million Chinese social media instances, evenly split between posts and comments. The corresponding SFT model generates three candidate translations per instance, which are scored by the Agent-enhanced Judger. An instance is considered hard if at least two of its three translations receive a score of 0 or 1. This criterion selects inputs with repeated low-scoring translations rather than relying on a single unsuccessful generation. We then uniformly sample 40K hard instances for GRPO training at each model scale.

\paragraph{Translation GRPO.}

The translation model receives a reward combining output format, reasoning length, and cultural effectiveness:
\begin{equation*}
r
=
r_{\mathrm{format}}
+
r_{\mathrm{length}}
+
r_{\mathrm{culture}}.
\end{equation*}

The format reward encourages the model to retain the CoT--translation structure learned during SFT, with contextual and cultural analysis preceding the final translation:
\begin{equation*}
r_{\mathrm{format}} =
\begin{cases}
0.5, & \text{output CoT-translation format},\\
0, & \text{otherwise}.
\end{cases}
\end{equation*}

Since the CoT is intended to summarize translation-relevant considerations rather than provide an extended analysis, we use a length reward to discourage unnecessarily verbose reasoning:
\begin{equation*}
r_{\mathrm{length}} =
\begin{cases}
0.5, & L_{\mathrm{reason}} \leq 1024,\\
0, & \text{otherwise},
\end{cases}
\end{equation*}
where $L_{\mathrm{reason}}$ is the number of reasoning tokens.

The cultural-effectiveness reward uses the score $s$ assigned by the Agent-enhanced Judger. To align the reward with the acceptance criterion used in data refinement and hard-example mining, we assign the same reward to scores 2 and 3, while giving only limited credit to score 1:
\begin{equation*}
r_{\mathrm{culture}} =
\begin{cases}
0,   & s=0,\\
0.1, & s=1,\\
1.0, & s\in\{2,3\}.
\end{cases}
\end{equation*}
This mapping prioritizes culturally acceptable translations rather than distinguishing between quality levels within the acceptable range.

\section{Experiments Setting}

\subsection{Translation Model Setup}
\label{sec:exp-setup}

\paragraph{Models and Baselines.}
\label{sec:baselines}

We evaluate CUE-MT at the 9B, 27B, and 35B-A3B scales based on Qwen3.5~\citep{qwen35blog}. For each scale, we compare the original base model, the supervised fine-tuned model (\textbf{CUE-MT w/o GRPO}), and the final CUE-MT model trained with Agent-enhanced rewards. We include two dedicated translation model families as primary baselines. \textbf{Seed-X}~\citep{cheng2025seedx} is a multilingual translation model trained with chain-of-thought supervised fine-tuning and further optimized with reinforcement learning; we report both its SFT and PPO variants. \textbf{Hy-MT2}~\citep{zheng2026hymt2familyfastefficient} is a family of multilingual translation models designed for complex real-world translation scenarios; we evaluate its 7B and 30B-A3B variants. We further compare against strong general LLMs, including \textbf{Gemini-3.1-Pro}~\citep{google-gemini-3}, \textbf{Gemini-3.7-Flash}~\citep{gemini-37-flash}, GLM-5.2~\citep{glm5team2026glm5vibecodingagentic}, \textbf{Qwen3.8-Max}~\citep{qwen38}, \textbf{GPT-5.6}~\citep{gpt-5d6}, and \textbf{MiniMax-M3}~\citep{MiniMax}. Together, these baselines cover specialized translation systems and general LLMs across a broad range of model scales. Detailed model configurations are provided in Appendix~\ref{app:training-details}.

\paragraph{Evaluation Benchmarks.}

We construct \textbf{CUE-MT-Bench} to evaluate social media translation. It contains titles, posts, and contextualized comments collected from publicly available social-media data. We initially collect 1,500 instances and remove examples requiring unavailable visual or external information or exceeding 200 Chinese characters. The resulting benchmark contains 980 instances, including 341 titles, 312 posts, and 327 comments. For contextualized comments, the corresponding title and post or preceding comment are provided as context. CUE-MT-Bench was sampled during April 30--May 30, 2026, while the sampling cutoff for Judger training and translation-model training data is November 30, 2025 (Appendix~\ref{app:temporal-scope}). We additionally evaluate translation performance on \textbf{CULTURE-MT}~\citep{CULTURE-MT}, \textbf{RedTrans}~\citep{guo2025redefiningmachinetranslationsocial}, \textbf{Seed-X-Challenge}~\citep{cheng2025seedx}, and \textbf{FLORES-zh2en}~\citep{goyal2021flores101evaluationbenchmarklowresource}.

\paragraph{Translation Evaluation.}

On CUE-MT-Bench, CULTURE-MT, RedTrans, and Seed-X-Challenge, we report cultural-effectiveness evaluation using two LLM-based evaluators: the GRPO-backed Agent-enhanced Judger and the Gemini API Judger (Appendix~\ref{app:llm-evaluation}). For each evaluator, cultural effectiveness is measured as the percentage of translations assigned a score of 2 or 3 under the four-level rubric of CULTURE-MT~\citep{CULTURE-MT}. On FLORES-zh2en, we additionally report BLEU~\citep{papineni2002bleu}, chrF~\citep{popovic2015chrf}, and COMET~\citep{rei2020comet}. All benchmark evaluations use greedy decoding. We additionally conduct human evaluation on CUE-MT-Bench.

\subsection{Judger Setup}

\paragraph{Models and Baselines.}

Our cultural-effectiveness Judger is initialized from Qwen3.5-27B. We evaluate both the static Judger without retrieval and the Agent-enhanced Judger augmented with external evidence. The retrieval Agent uses GLM-5.2~\citep{glm5team2026glm5vibecodingagentic} for query planning and hint generation, while external retrieval is performed using the GLM-4.7-based Zhipu search service\footnote{\url{https://modelscope.cn/mcp/servers/ZhipuAI/Zhipu-Web-Search}}.

We compare against the corresponding \textbf{static Judgers}, the \textbf{Gemini API Judger} based on Gemini-3.1-Pro, and our reimplementation of \textbf{Static Judger (Hint GRPO)}~\citep{qiu2026metricsrewardworsttranslations}. Detailed baseline and ablation settings are provided in Appendix~\ref{app:judger-baselines}.

The retrieval ablation keeps the GRPO-aligned Judger, evaluation inputs, scoring pipeline, and query-trigger decisions unchanged, while removing external evidence and generating hints without retrieval. This isolates the contribution of retrieved cultural evidence beyond hint generation.

\begin{table*}[t]
\small
\centering
\caption{
Evaluation of cultural-effectiveness Judgers on our human-annotated benchmark.
}
\label{tab:judger-main}
\resizebox{0.9\textwidth}{!}{\begin{tabular}{lcccc}
\toprule
\textbf{Evaluator}
& \textbf{Bin. Acc. (\%)}
& \textbf{$\kappa$}
& \textbf{4-way Acc. (\%)}
& \textbf{Spearman} \\
\midrule
Gemini API Judger \citep{google-gemini-3}
& 75.21 & 0.5043 & 48.07 & 0.6059 \\

Static Judger (SFT)  \citep{CULTURE-MT}
& 71.14 & 0.4229 & 38.57 & 0.4313 \\

Static Judger (SFT + GRPO)
& 70.86 & 0.4171 & 50.00 & 0.5826 \\

Static Judger (Hint GRPO baseline) \citep{qiu2026metricsrewardworsttranslations}
& 69.71 & 0.3944 & 47.64 & 0.5646 \\

\midrule

Agent-enhanced Judger (GRPO backbone, w/o retrieval)
&74.00&0.4802&44.50&0.5523\\

Agent-enhanced Judger (SFT backbone)
&76.64&0.5329&51.21&0.6279\\

Agent-enhanced Judger (GRPO backbone)
&\textbf{77.86}&\textbf{0.5571}&\textbf{52.57}&\textbf{0.6558} \\
\bottomrule
\end{tabular}}
\end{table*}

\paragraph{Evaluation Benchmark.}

We construct \textbf{CUE-Judge-Bench} to evaluate cultural-effectiveness Judgers. From 636 source instances sampled from CUE-MT-Bench, we collect candidate translations generated by 28 model variants, including external baselines and our own CUE-MT variants. These variants cover general LLMs, dedicated translation models, and base, SFT, and GRPO-trained models. All translations are manually annotated using the four-level cultural-effectiveness scale $\{0,1,2,3\}$ following the annotation procedure described in Appendix~\ref{app:human-evaluation}. We then sample 350 examples from each score level, yielding a balanced benchmark of 1,400 source--translation pairs.

\paragraph{Judger Evaluation.}

We evaluate Judger predictions against adjudicated human annotations using binary accuracy, Cohen's $\kappa$, four-way accuracy, and Spearman correlation. Binary accuracy and $\kappa$ compare scores 0--1 against 2--3, while four-way accuracy and Spearman correlation evaluate agreement over the full four-level scale. All Judger evaluations use greedy decoding.

\section{Experiments Results}

\subsection{Retrieval-Enhanced Cultural-Effectiveness Judging}
\label{sec:judger-results}

We examine the effect of Judger optimization and evaluate whether retrieval-based cultural evidence improves agreement with human judgments.

\paragraph{GRPO improves fine-grained cultural score calibration.} Compared with the SFT checkpoint, Static Judger (SFT + GRPO) substantially improves fine-grained agreement with human annotations. Four-way accuracy increases by 11.43 percentage points (from 38.57\% to 50.00\%), and Spearman correlation improves by 0.1513 (from 0.4313 to 0.5826). Meanwhile, binary accuracy and Cohen's $\kappa$ remain slightly lower, suggesting that GRPO primarily improves score calibration across four cultural-effectiveness levels rather than coarse binary acceptance decisions.

\paragraph{Retrieval provides additional gains beyond parameterized knowledge.} Adding Agent-based retrieval further improves Judger performance. With the SFT backbone, Agent-enhanced Judger improves binary accuracy by 5.50 percentage points, four-way accuracy by 12.64 percentage points, and Spearman correlation by 0.1966 compared with the static SFT Judger. Starting from the GRPO backbone, Agent-enhanced Judger achieves the best overall performance, reaching 77.86\% binary accuracy, 0.5571 Cohen's $\kappa$, 52.57\% four-way accuracy, and 0.6558 Spearman correlation.

\paragraph{External evidence contributes beyond generated hints.} The no-retrieval ablation shows that removing external evidence substantially degrades Judger performance. Four-way accuracy drops by 8.07 percentage points (from 52.57\% to 44.50\%), while binary accuracy, Cohen's $\kappa$, and Spearman correlation also decrease. This comparison demonstrates that retrieval provides useful cultural evidence beyond GLM-5.2-generated hints alone, while not attributing the improvement specifically to newly emerging knowledge. Qualitative examples further illustrate improved contextual interpretation and remaining challenges in error coverage and severity calibration (Appendix~\ref{app:judger-error-analysis}).
\begin{table*}[t]
\centering
\small
\caption{
Comparison of CUE-MT with representative translation models on cultural-effectiveness benchmarks.
Agent, Gemini, and Human denote evaluations by the GRPO-backed Agent-enhanced Judger, Gemini API Judger, and human annotators, respectively.
Best results among comparable-size models are shown in bold.
}
\label{tab:main_results}

\resizebox{0.9\textwidth}{!}{
\begin{tabular}{lccccccccc}
\toprule
& \multicolumn{3}{c}{\textbf{CUE-MT-Bench}}
& \multicolumn{2}{c}{\textbf{CULTURE-MT}}
& \multicolumn{2}{c}{\textbf{RedTrans}}
& \multicolumn{2}{c}{\textbf{Seed-X-Challenge}} \\
\cmidrule(lr){2-4}
\cmidrule(lr){5-6}
\cmidrule(lr){7-8}
\cmidrule(lr){9-10}

\textbf{Model}
& \textbf{Agent}
& \textbf{Gemini}
& \textbf{Human}
& \textbf{Agent}
& \textbf{Gemini}
& \textbf{Agent}
& \textbf{Gemini}
& \textbf{Agent}
& \textbf{Gemini}
\\
\midrule

\rowcolor{gray!12}
\multicolumn{10}{l}{\textit{Model scale: $\leq$9B}} \\

Seed-X-7B-SFT
&67.96&64.75&77.73
&72.55&52.76
&78.31&76.24
&82.23&76.53
\\

Seed-X-7B-PPO
&72.01&71.46&79.78
&80.24&69.17
&80.26&80.82
&84.77&78.68
\\

HY-MT2-7B
&68.13&67.14&77.02
&79.14&74.45
&79.18&80.10
&87.82&86.29
\\

Qwen3.5-9B
&67.28&56.73&68.74
&81.14&75.15
&80.65&84.13
&89.34&85.28
\\

CUE-MT w/o GRPO
&80.08&90.82&90.09
&92.64&93.71
&86.53&90.86
&89.85&87.82
\\

\rowcolor{cueblue}
\textbf{CUE-MT}
&\textbf{83.04}
&\textbf{93.06}
&\textbf{90.71}
&\textbf{94.31}
&\textbf{94.09}
&\textbf{87.61}
&\textbf{91.67}
&\textbf{93.40}
&\textbf{92.89}
\\

\midrule

\rowcolor{gray!12}
\multicolumn{10}{l}{\textit{Model scale: 30--35B-A3B}} \\

HY-MT2-30B-A3B
&73.65&79.37&83.67
&86.63&87.89
&85.05&87.61
&91.88&88.32
\\

Qwen3.5-35B-A3B
&77.53&58.63&75.18
&85.93&85.54
&85.54&89.57
&93.40&92.39
\\

CUE-MT w/o GRPO
&86.21&93.78&87.33
&88.72&88.82
&88.27&93.74
&92.39&\textbf{95.43}
\\

\rowcolor{cueblue}
\textbf{CUE-MT}
&\textbf{87.24}
&\textbf{94.90}
&\textbf{91.11}
&\textbf{91.02}
&\textbf{90.76}
&\textbf{90.23}
&\textbf{94.54}
&\textbf{94.42}
&94.92
\\

\midrule

\rowcolor{gray!12}
\multicolumn{10}{l}{\textit{Model scale: 27B}} \\

Qwen3.5-27B
&79.88&61.80&77.53
&91.52&93.29
&87.12&91.77
&93.40&91.88
\\

CUE-MT w/o GRPO
&84.08&93.88&90.31
&95.31&\textbf{96.30}
&89.67&95.48
&\textbf{94.42}&93.40
\\

\rowcolor{cueblue}
\textbf{CUE-MT}
&\textbf{88.27}
&\textbf{95.61}
&\textbf{92.65}
&\textbf{96.31}
&96.20
&\textbf{90.13}
&\textbf{95.54}
&\textbf{94.42}
&\textbf{96.45}
\\

\bottomrule
\end{tabular}
}
\end{table*}

\subsection{Main Results on Social Media Translation}
\label{sec:main-results}

Table~\ref{tab:main_results} reports CUE-MT-Bench results using human evaluation as the primary assessment, with Agent-enhanced Judger and Gemini judgments as scalable automatic evaluators.

\paragraph{Supervised training provides the major performance improvement.} Across all three model scales, CUE-MT w/o GRPO substantially improves over the corresponding Qwen3.5 base models. On human evaluation, supervised training increases cultural effectiveness by 21.35 percentage points at 9B (68.74\% $\rightarrow$ 90.09\%), 12.78 percentage points at 27B (77.53\% $\rightarrow$ 90.31\%), and 12.15 percentage points at 35B-A3B (75.18\% $\rightarrow$ 87.33\%). Similar improvements are observed under Agent-enhanced Judger and Gemini evaluation, indicating that the gains are not specific to a single evaluator.

\paragraph{GRPO further improves translation quality after supervised alignment.} Starting from the supervised models, GRPO provides additional improvements on culturally challenging social media translation. At 27B, GRPO improves Agent evaluation by 4.19 percentage points and human evaluation by 2.34 percentage points. At 35B-A3B, human-rated cultural effectiveness increases by 3.78 percentage points. The 9B model also shows consistent gains across human and automatic evaluations, demonstrating that Agent-enhanced rewards provide useful optimization signals beyond supervised cultural alignment.

\paragraph{CUE-MT achieves strong performance across model scales.} CUE-MT achieves the strongest overall results among comparable open-source translation models at all evaluated scales. At 9B, CUE-MT reaches 90.71\% human-rated cultural effectiveness, improving over the strongest specialized baseline Seed-X-7B-PPO by 10.93 percentage points. At 27B, CUE-MT achieves 92.65\%, outperforming Qwen3.5-27B by 15.12 percentage points. At 35B-A3B, CUE-MT reaches 91.11\%, exceeding HY-MT2-30B-A3B by 7.44 percentage points. These results demonstrate that cultural-effectiveness-guided training enables smaller and medium-scale models to achieve competitive performance with substantially larger systems.

\begin{figure}[h]
    \centering
    \includegraphics[width=0.48\textwidth]{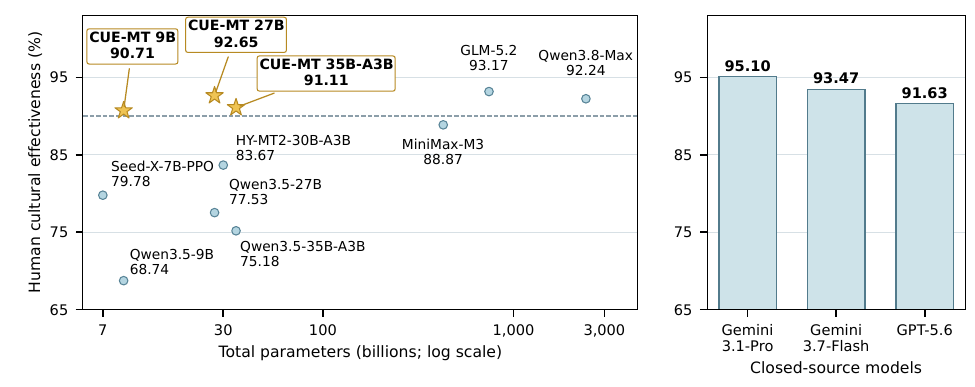}
    \caption{{Human-rated cultural effectiveness on CUE-MT-Bench (percentage scored 2 or 3). CUE-MT combines compact model size with scores approaching those of larger systems.}}
    \label{fig:main_large}
\end{figure}

\paragraph{Performance transfers to public benchmarks.} Beyond CUE-MT-Bench, CUE-MT consistently improves over corresponding base models on CULTURE-MT, RedTrans, and Seed-X-Challenge. Under Agent evaluation, CUE-MT improves over the corresponding Qwen3.5 base models by 13.17, 6.96, and 4.06 percentage points at 9B, with consistent gains at 27B and 35B-A3B. These improvements across diverse benchmarks demonstrate that cultural-effectiveness-guided training generalizes beyond the proposed benchmark.

\begin{table}[t]
\centering
\caption{
Ablation study on CUE-MT-27B.
All GRPO variants start from the same CUE-MT w/o GRPO checkpoint.
Hard-example mining and reward evaluator effects are evaluated under four combinations.
}
\label{tab:ablation}
\small

\resizebox{0.48\textwidth}{!}{
\begin{tabular}{lccc}
\toprule
\textbf{Training Strategy}
& \textbf{Agent}
& \textbf{Gemini}
& \textbf{Human}\\
\midrule

Qwen3.5-27B
&79.88&61.80&77.53\\

CUE-MT w/o GRPO
&84.08&93.88&90.31\\

\midrule
\rowcolor{gray!12}\multicolumn{4}{l}{\textit{CUE-MT-27B}} \\
Hard Mining + Agent Reward
&\textbf{88.27}
&\textbf{95.61}
&\textbf{92.65}\\

Hard Mining + Static Reward
&84.08
&94.59
&91.32\\

Random Data + Agent Reward
&85.70
&95.51
&92.04\\

Random Data + Static Reward
&84.48
&94.69
&91.63\\

\bottomrule
\end{tabular}}
\end{table}

\subsection{Ablation Studies}
\label{sec:ablation}
We conduct an ablation study on CUE-MT-27B to analyze the contributions of hard-example mining and Agent-enhanced rewards. All GRPO variants start from the same CUE-MT w/o GRPO checkpoint. We compare two data-selection strategies (hard-example mining and random sampling) and two reward settings (Agent-enhanced Judger and static Judger). For hard-example mining, the corresponding Judger is used to identify difficult instances and provide rewards during GRPO training. Random selection samples 40K instances without replacement from the same 500K candidate pool used for mining (seed 42), while all other training configurations remain fixed.

\paragraph{Agent-enhanced Judger provides more effective training signals.} Replacing the static Judger with the Agent-enhanced Judger consistently improves GRPO performance under both data-selection strategies. With hard-example mining, Agent-enhanced rewards improve human evaluation by 1.33 percentage points, compared with 0.41 percentage points under random sampling. Similar improvements are observed under Agent-enhanced Judger evaluation (4.19 vs. 1.22 percentage points) and Gemini evaluation (1.02 vs. 0.82 percentage points). These results demonstrate that retrieval-enhanced cultural evaluation provides more effective optimization signals for GRPO training.

\paragraph{Hard-example mining benefits from improved evaluation signals.} Using Agent-enhanced rewards, hard-example mining improves human evaluation by 0.61 percentage points over random sampling, with additional gains of 2.57 and 0.10 percentage points under Agent-enhanced Judger and Gemini evaluation, respectively. In contrast, hard-example mining with the static Judger provides limited improvements and slightly decreases performance under all three evaluations. These results suggest that a stronger Judger improves not only the reward quality but also the selection of challenging examples.

These results show that retrieval-enhanced evaluation and difficulty-aware data selection are complementary: the Agent-enhanced Judger improves the quality of optimization signals, while hard-example mining focuses training on more informative examples.

\begin{figure}[t]
\centering
\includegraphics[width=0.48\textwidth]{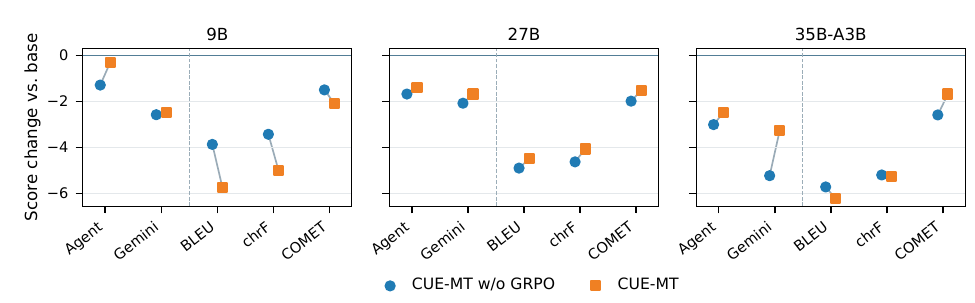}
\caption{{
Changes in FLORES-zh2en scores relative to the corresponding Qwen3.5 base models.
}}
\label{fig:flores}
\end{figure}

\subsection{Impact on General-Domain Translation}
\label{sec:flores-results}

We evaluate FLORES-zh2en to examine whether social-media specialization affects general-domain translation capability. Figure~\ref{fig:flores} reports score changes relative to the corresponding Qwen3.5 base models, with full results provided in Table~\ref{tab:flores}.

Social-media specialization leads to decreases on reference-based metrics, particularly BLEU and chrF, while CUE-MT maintains Agent and Gemini scores above 93\% across all three model scales. This discrepancy suggests that the observed metric changes partly reflect a shift toward more natural and conversational translations rather than uniform degradation. For example, ``assisted Carter in diplomatically handling world affairs'' is translated as ``helped Carter handle a lot of international stuff.'' The translation adopts a more conversational style, which reduces reference overlap, but it also weakens the diplomatic nuance carried by the original expression. Such cases show that social-media specialization introduces a trade-off between conversational adaptation and fine-grained fidelity. We also observe a small number of genuine semantic errors, indicating room for further improvement. Overall, CUE-MT maintains general-domain translation capability while adapting its generation style toward social-media-oriented translation.

\section{Conclusion}

We propose \textbf{CUE-MT}, a social media translation framework that enhances cultural-effectiveness evaluation with input-dependent external evidence. By using retrieval-enhanced Judger feedback for data refinement, hard-example mining, and reinforcement learning, CUE-MT improves cultural effectiveness across multiple model scales and benchmarks. Our results show that external evidence provides more reliable training signals than learned evaluation knowledge alone, especially when combined with difficulty-aware data selection. Future work will investigate robustness to noisy evidence and broader cultural and multilingual scenarios.

\section*{Limitations}
Our work has several limitations. First, the effectiveness of retrieval-enhanced evaluation depends on the quality and availability of external evidence. Although retrieval improves Judger performance in our experiments, noisy, incomplete, or misleading evidence may introduce additional uncertainty into the evaluation process.

Second, our cultural-effectiveness evaluation focuses on Chinese-to-English social media translation and relies on a four-level human annotation rubric. While this setting enables controlled analysis of cultural knowledge gaps, broader cultural phenomena, language pairs, and more fine-grained evaluation criteria remain to be explored.

Third, our experiments are conducted on a limited set of model scales and backbone families. Although CUE-MT demonstrates consistent improvements across the evaluated Qwen3.5-based models, the general applicability of retrieval-enhanced cultural evaluation to other architectures and larger multilingual systems requires further investigation.

\section*{Ethical Considerations}

We use publicly available social-media data collected in accordance with applicable crawling requirements. To reduce privacy risks, we anonymized explicit identifiers, including usernames and personal names, removed identifiable information where applicable, and applied filtering to remove offensive and harmful content. Appendix~\ref{app:data-cleaning} describes the cleaning procedure. Human annotators were recruited through an outsourcing company and were based in China and Malaysia. They provided informed consent and were compensated according to the number of annotated items. The annotation task focused on evaluating translation quality and cultural effectiveness using publicly available text processed through our data-cleaning pipeline. Our models aim to improve the preservation of cultural meaning in social-media translation. However, cultural judgments may vary across communities, and model-based evaluation may reflect particular linguistic or cultural perspectives. Translation errors and unreliable retrieved evidence may also misrepresent the original speaker's meaning. The models should therefore be used with appropriate human oversight, especially when translation errors may have consequential effects. We plan to release the trained models. AI assistance was used to correct English grammar and improve the readability, organization, and structure of the manuscript.
% \section*{Acknowledgments}

% This document has been adapted
% by Steven Bethard, Ryan Cotterell and Rui Yan
% from the instructions for earlier ACL and NAACL proceedings, including those for
% ACL 2019 by Douwe Kiela and Ivan Vuli\'{c},
% NAACL 2019 by Stephanie Lukin and Alla Roskovskaya,
% ACL 2018 by Shay Cohen, Kevin Gimpel, and Wei Lu,
% NAACL 2018 by Margaret Mitchell and Stephanie Lukin,
% Bib\TeX{} suggestions for (NA)ACL 2017/2018 from Jason Eisner,
% ACL 2017 by Dan Gildea and Min-Yen Kan,
% NAACL 2017 by Margaret Mitchell,
% ACL 2012 by Maggie Li and Michael White,
% ACL 2010 by Jing-Shin Chang and Philipp Koehn,
% ACL 2008 by Johanna D. Moore, Simone Teufel, James Allan, and Sadaoki Furui,
% ACL 2005 by Hwee Tou Ng and Kemal Oflazer,
% ACL 2002 by Eugene Charniak and Dekang Lin,
% and earlier ACL and EACL formats written by several people, including
% John Chen, Henry S. Thompson and Donald Walker.
% Additional elements were taken from the formatting instructions of the \emph{International Joint Conference on Artificial Intelligence} and the \emph{Conference on Computer Vision and Pattern Recognition}.


\bibliography{custom}

\appendix
\section{Static Judger Training}
\label{app:training-details-judger}

We initialize the static Judger from Qwen3.5-27B and train it through two SFT stages followed by GRPO.

\paragraph{Supervised Fine-Tuning.}
We first perform full-parameter SFT on the Gemini-annotated data, then continue SFT on the human-annotated data. Both stages use MS-SWIFT\footnote{\url{https://github.com/modelscope/ms-swift}} with DeepSpeed ZeRO-3\footnote{\url{https://github.com/deepspeedai/DeepSpeed}} in bfloat16 precision. We use a per-device batch size of 4, gradient accumulation over 2 steps, and a maximum sequence length of 4096 tokens. Each stage runs for three epochs with a learning rate of $5\times10^{-6}$. The warmup ratio is 0.1 in the first stage and 0 in the second.

\paragraph{GRPO Training.}
Starting from the second-stage SFT checkpoint, we further optimize the Judger with GRPO using Relax\footnote{\url{https://github.com/redai-studio/Relax}} on 32 GPUs across four nodes, with eight GPUs per node, in a colocated configuration. We use a tensor parallelism degree of 4, sequence parallelism, full activation recomputation, and optimizer-state offloading to the CPU.

For each prompt, we sample 8 responses at temperature 0.7 and top-$p$ 0.8, with a maximum response length of 2048 tokens. Each rollout batch contains 128 prompts, yielding 1024 responses. The global training batch size is 64 prompt--response pairs. We use the GRPO advantage estimator without a KL penalty or entropy regularization, with asymmetric clipping margins of $\epsilon_{\mathrm{low}}=0.2$ and $\epsilon_{\mathrm{high}}=0.28$. Optimization uses Adam with a constant learning rate of $1\times10^{-6}$, $\beta_1=0.9$, $\beta_2=0.98$, and weight decay of $0.1$. Training runs for 600 rollout steps.

\paragraph{Validation and Checkpoint Selection.}
We reserve 5\% of the Judger GRPO data pool for validation and exclude these examples from all GRPO updates. We evaluate every 10 rollout steps using greedy decoding and use binary accuracy as the primary checkpoint-selection metric.

% \paragraph{Hint GRPO Baseline.}
% {We reimplement the Hint GRPO approach of \citet{qiu2026metricsrewardworsttranslations} on Qwen3.5-27B, starting directly from the base checkpoint rather than our two-stage SFT Judger. It uses the same 20K human-labeled difficult-example pool as our Judger GRPO, and the hint is generated by GLM5.2. Thus, its initialization differs from our static SFT+GRPO Judger; the no-retrieval Agent ablation provides the matched comparison with a fixed GRPO backbone.}

% \paragraph{No-Retrieval Agent Ablation.}
% {The ablation in Table~\ref{tab:judger-main} retains the GRPO-aligned Judger, evaluation inputs, scoring pipeline, and the complete Agent's original query-trigger decisions. External retrieval is disabled. Only the same 52.64\% of inputs that originally triggered retrieval receive GLM-5.2-generated judging points, produced without retrieved evidence; all other inputs receive no hint. This preserves the trigger population when comparing evidence-conditioned and evidence-free hints. Enabling retrieval increases binary and four-way accuracy by 3.86 and 8.07 percentage points, respectively, with gains of 0.0769 in $\kappa$ and 0.1035 in Spearman correlation. Hints alone do not uniformly improve the static GRPO Judger: they raise binary accuracy but reduce four-way accuracy from 50.00\% to 44.50\%. These results distinguish retrieval from hint generation in this configuration; they do not establish statistical significance or attribute the gains specifically to newer knowledge.}

\section{Design and Prompt of Retrieval Agent}
\subsection{Retrieval Execution}
\label{app:retrieval-execution}

GLM-5.2 performs query planning and hint generation. We use the Zhipu retrieval API with GLM-4.7 to retrieve the top five results per query. Retrieved evidence and generated hints are cached for reuse. If the planner generates no queries, the Judger evaluates the translation without hints. If retrieval returns no useful evidence or the API call fails, GLM-5.2 generates concise general judging guidance without external evidence, following the hint-generation prompt. On CUE-Judge-Bench, the planner generates at least one query for 52.64\% of the 1,400 source--translation pairs. 
%We define this proportion as the retrieval trigger rate, which is distinct from the actual API-call frequency after caching and the retrieval success rate.

\subsection{Prompts for the Agent-enhanced Judger}
\label{app:agent-prompts}

This appendix provides the full prompts used by the search planner $g_{\mathrm{plan}}$ (Appendix~\ref{app:prompt-plan}) and the hint generator $g_{\mathrm{hint}}$ (Appendix~\ref{app:prompt-hint}), together with the hint injection format used by the Judger (Appendix~\ref{app:hint-injection}). The Judger scoring prompt is inherited from prior work; we refer readers to \cite{CULTURE-MT} for its full content. The prompts are presented in English for readability.

% ---------------------------------------------------------------
\subsubsection{Search Planning Prompt ($g_{\mathrm{plan}}$)}
\label{app:prompt-plan}

The search planner determines whether external knowledge is required
and generates a small set of searchable queries when necessary.

\paragraph{System prompt.}

\begin{promptbox}[System Prompt for Search Planning]
You are a pre-analysis assistant for translation evaluation.

Given a Chinese source text, with optional context, and its English
translation, determine whether external web search is required to
correctly evaluate the translation.

Return JSON only. Do not use Markdown.

\{
  ``needs\_search'': [``query1'', ``query2''],
  ``reason'': ``brief explanation''
\}

Rules:

1. Search only when the input contains internet memes, fandom or anime
jargon, Xiaohongshu platform slang, names of brands, works, people, or
places, or culturally loaded expressions that are likely to be
mistranslated literally.

2. Each query must be short and independently searchable.

3. Chinese source expressions must remain in Chinese and must not be
translated into English in the query.

4. When an official English alias is required, use the format
``Chinese term English alias''. Do not use the English alias alone.

5. Do not search for ordinary words, ordinary sentences, or
unambiguous grammatical errors.

6. Generate at most four queries.

7. Return an empty list when no external search is required.
\end{promptbox}

\paragraph{User template.}

\begin{promptbox}[User Template for Search Planning]
[Source / Context]

\{source\_input\}

[Translation]

\{translation\}
\end{promptbox}

% ---------------------------------------------------------------
\subsubsection{Hint Generation Prompt ($g_{\mathrm{hint}}$)}
\label{app:prompt-hint}

The hint generator filters the retrieved information and converts it
into concise, translation-specific judging points. It does not assign
the final cultural-effectiveness score.

\paragraph{System prompt.}

\begin{promptbox}[System Prompt for Hint Generation]
You are a rigorous but restrained translation-quality evaluation
assistant.

Your task is not to assign a final score. Instead, provide referenceable judging points for a downstream Judger.

Based on the source text, the candidate translation, and the search
results, output JSON in the following format:

\{
  ``judging\_points'': [``point 1'', ``point 2''],
  ``risk\_flags'': [``risk label 1'', ``risk label 2'']
\}

Requirements:

1. Judging points should describe how the translation should be evaluated. Relevant aspects include missing or mistranslated content, the correctness of proper nouns, the preservation of placeholders and structural tags, and whether the register is appropriate for the context.

2. Search results are auxiliary information only. If they are irrelevant or unreliable, ignore them explicitly. Do not fabricate knowledge.

3. You may identify possible translation problems, but you must not assign a final score of 0, 1, 2, or 3.

4. For Chinese internet memes, fandom expressions, special forms of address, and titles of works, explain the intended meaning and the risk of literal translation.

5. If no useful external knowledge is available, return a small number of general judging points.

6. Generate at most six judging points, and keep each point concise.
\end{promptbox}

\paragraph{User template.}

\begin{promptbox}[User Template for Hint Generation]
[Source / Context]

\{source\_input\}

[Translation]

\{translation\}

[Search Queries]

\{terms\}

[Search Results]

\{search\_results\}
\end{promptbox}

% ---------------------------------------------------------------
\subsubsection{Hint Injection Format}
\label{app:hint-injection}

The structured hint object $H$ is serialized into an auxiliary text block and inserted before the source--translation pair in the Judger prompt. The instruction explicitly states that the hints are supplementary and should not override evidence from the source text and candidate translation.

\begin{promptbox}[Hint Injection Template]
Referenceable judging points:

These points are auxiliary information only. Rely primarily on the
source text and the candidate translation, and do not adopt the hints
blindly.

1. \{judging\_point\_1\}

2. \{judging\_point\_2\}

...

Possible risk flags:

\{risk\_flag\_1\}, \{risk\_flag\_2\}, ...

Please evaluate the translation based on the following source or
context and candidate translation.

[Source / Context]

\{source\_input\}

[Translation]

\{translation\}
\end{promptbox}

% ===============================================================
% Translation Model Training Prompts
% ===============================================================

\section{Prompts for Translation Model Training}
\label{app:translation-training-prompts}

This appendix presents the prompt templates used in our translation
model training pipeline. All prompts are translated into English for
clarity. Variables enclosed in braces are replaced with the
corresponding sample content during data construction.

\subsection{Prompt for Translation Self-Refinement}
\label{app:translation-refinement}

{For translations that do not pass the initial static-Judger assessment, we perform one round of self-refinement. The model receives the source text, optional context, its initial translation, and feedback from the static Judger; retrieval is not used in this stage.}

\begin{promptbox}[Prompt for Translation Self-Refinement]
You are an expert in social media translation. Please revise the
initial English translation according to the source text, the
available context, and the evaluation feedback.

The revised translation should preserve the original meaning,
cultural references, emotional tone, and intended pragmatic effect.
It should also be natural and understandable to English-speaking
social media users.

Follow these requirements:

1. Carefully address the translation problems identified in the
evaluation feedback.

2. Do not change parts of the translation that are already correct
unless doing so improves consistency or fluency.

3. Placeholders in the form ``\#placeholder-x\#'' must not be
translated. Preserve each placeholder at the corresponding position.

4. Preserve structural tags such as
\texttt{<title></title>},
\texttt{<content></content>}, and
\texttt{<comment></comment>}.

5. Output only the revised translation. Do not explain your revisions
or provide any additional content.

Context:

\{context\}

Source text:

\{source\}

Initial translation:

\{translation\}

Judger feedback:

\{feedback\}

Please provide the revised translation:
\end{promptbox}

% ---------------------------------------------------------------
\subsection{Prompts for Initial Translation Generation}
\label{app:translation-prompts}

We use two prompts to generate the initial English translations.
The first prompt is used for titles and posts, while the second is
used for comments that require a post or parent comment as context.

% ---------------------------------------------------------------
\subsubsection{Title and Post Translation Prompt}
\label{app:title-post-prompt}

\begin{promptbox}[Prompt for Title and Post Translation]
You are an expert in social media translation. Please translate the
following content into English. Follow these requirements:

1. Placeholders in the form ``\#placeholder-x\#'' must not be
translated. Preserve each placeholder at the corresponding position
in the translated text.

2. Output only the translated result. Do not provide explanations or
any additional content.

3. Preserve structural tags such as
\texttt{<title></title>},
\texttt{<content></content>}, and
\texttt{<comment></comment>}.
The translated result must follow the same tag structure as the input.

Content to be translated:

\{input\}

Please provide the translated result:
\end{promptbox}

% ---------------------------------------------------------------
\subsubsection{Contextualized Comment Translation Prompt}
\label{app:comment-translation-prompt}

\begin{promptbox}[Prompt for Contextualized Comment Translation]
You are an expert in social media translation. Please translate the
following content into English. Follow these requirements:

1. Placeholders in the form ``\#placeholder-x\#'' must not be
translated. Preserve each placeholder at the corresponding position
in the translated text.

2. Output only the translated result. Do not provide explanations or
any additional content.

3. Preserve structural tags such as
\texttt{<title></title>},
\texttt{<content></content>}, and
\texttt{<comment></comment>}.
The translated result must follow the same tag structure as the input.

When the sentence to be translated contains an unclear reference or
an ambiguous meaning, use the following post or parent-comment context
to determine its intended meaning:

\{context\}

Return only the final translation and no other content.

Sentence to be translated:

\{input\}

Please provide the translated result:
\end{promptbox}

% ---------------------------------------------------------------

% ---------------------------------------------------------------
\subsection{Prompt for Reasoning Trace Generation}
\label{app:cot-generation}

For each retained source--translation pair, we use GLM-5 to generate
a reasoning trace consisting of key contextual cues and translation
points. The model identifies culturally or semantically important
expressions in the source and explains how they should be understood
to reproduce the reference translation.

\begin{promptbox}[Prompt for Reasoning Trace Generation]
You are a multilingual translation expert for Xiaohongshu-style social
media content.

Given the context, source text, and reference translation, identify the
key translation points that are necessary to produce a translation
like the reference.

First, identify important words or phrases directly from the
\textbf{source text}, and then briefly explain their meanings or the
main considerations for translating them.

Keep the analysis concise and direct.

Example:

Context:
None

Source:
\begin{CJK*}{UTF8}{gbsn}
真的是自嗨的剧情，不好看。 这不是高智商，这是走到哪里都有神助的玛丽苏
\end{CJK*}

Translation:
The plot is truly self-indulgent and unengaging; this isn't a display
of high intelligence but a Mary Sue story where the protagonist is
divinely favored and effortlessly succeeds wherever she goes.

Translation Points:

Pay attention to the following points when translating:

1. ``\begin{CJK*}{UTF8}{gbsn}自嗨\end{CJK*}'' refers to a plot that is immersed in the creator's own
self-satisfaction while ignoring the audience's perspective; it can be
rendered as ``self-indulgent plot.''

2. ``\begin{CJK*}{UTF8}{gbsn}高智商\end{CJK*}'' here refers to either a plot that requires logical
reasoning from the audience or characters whose actions demonstrate
intelligence.

3. ``\begin{CJK*}{UTF8}{gbsn}神助\end{CJK*}'' refers to the protagonist repeatedly receiving convenient
external help whenever difficulties arise.

4. ``\begin{CJK*}{UTF8}{gbsn}玛丽苏\end{CJK*}'' comes from the literary term ``Mary Sue'' and refers to
an unrealistically favored protagonist who seems to have a built-in
advantage or halo.

Context:
\{context\}

Source:
\{content\_zh\}

Translation:
\{trans\_en\}

Translation Points:
\end{promptbox}
% ---------------------------------------------------------------

\section{CUE-MT Training Details}
\label{app:training-details}

\paragraph{Data Collection and Validation Split.}

The 300K Judger-training pairs and the 3 million Chinese instances used for translation-model training are collected from publicly available online data before November 30, 2025. The two source collections do not overlap. Static-Judger refinement produces 2.4 million high-quality translation pairs, from which we reserve 1,200 examples for translation-model validation.

For translation-model training, the original 3 million Chinese instances are used for supervised data construction and later hard-example mining. CUE-Judge-Bench is constructed separately from CUE-MT-Bench source texts and therefore shares source instances with the translation benchmark by design.

\paragraph{Candidate Generation for Hard-Example Mining.}

After supervised fine-tuning, we perform hard-example mining separately for the 9B, 27B, and 35B-A3B models. For each model scale, we randomly sample 500K source instances from the original 3 million-instance translation-data pool, evenly split between posts and comments.

For each source instance, the corresponding SFT model generates three candidate translations using the same prompt, temperature 0.7, top-$p$ 0.8, random seed 42, and a maximum generation length of 4,096 tokens. The three candidates are scored by the Agent-enhanced Judger. An instance is selected as a hard example if at least two of its three candidates receive cultural-effectiveness scores of 0 or 1.

Sampling multiple candidates allows us to identify inputs where the model consistently struggles with cultural effectiveness rather than selecting examples based on a single unsuccessful generation. We uniformly sample 40K hard examples for GRPO training at each model scale.

\paragraph{Controlled Hard-Example Selection Ablation.}

For the data-selection ablation in Table~\ref{tab:ablation}, the random-selection variant samples 40K instances uniformly without replacement from the same 500K candidate pool used for hard-example mining, with random seed 42.

Both the hard-example and random-selection variants use the same number of training instances, the same GRPO training configuration, and the same training-step budget. Therefore, the comparison isolates the effect of difficulty-aware data selection.

The static reward baseline in this experiment uses the SFT+GRPO Judger rather than the SFT-only Judger, ensuring that the reward model quality remains consistent across the comparison.

\paragraph{Supervised Fine-Tuning.}

Supervised fine-tuning uses the structured chain-of-thought (CoT) targets described in Section~\ref{sec:training-pipeline}. The CoT consists of \emph{context analysis} and \emph{translation points}, which provide explicit supervision for understanding contextual dependencies, cultural expressions, and ambiguous meanings before generating the final translation.

We perform full-parameter supervised fine-tuning on Qwen3.5-9B, Qwen3.5-27B, and Qwen3.5-35B-A3B using MS-SWIFT. All models are trained for one epoch in bfloat16 precision with a learning rate of $1\times10^{-5}$, a warmup ratio of 0.05, and a maximum sequence length of 4,096 tokens.

The 9B and 27B models are trained on 16 GPUs (2 nodes $\times$ 8 GPUs) with DeepSpeed ZeRO-3, using a per-device batch size of 4. Gradient accumulation is set to 8 for Qwen3.5-9B and 4 for Qwen3.5-27B.

For the MoE-based Qwen3.5-35B-A3B model, we use tensor, pipeline, and expert parallelism with sizes 2, 4, and 2, respectively. We use a micro-batch size of 4 and a global batch size of 512, together with sequence parallelism and full activation recomputation. The MoE auxiliary-loss coefficient is set to $1\times10^{-6}$.

\paragraph{GRPO Training.}

Starting from the corresponding SFT checkpoints, we further optimize the three CUE-MT variants using GRPO within the Relax reinforcement-learning framework. Each model is trained on 32 GPUs (4 nodes $\times$ 8 GPUs) in a colocated setting, where training and rollout share the same GPUs. Optimizer states are offloaded to CPU to reduce GPU memory consumption.

Unless otherwise specified, all models use the same GRPO configuration. For each prompt, we sample 8 responses with temperature 0.7 and top-$p$ 0.8. The GRPO advantage estimator is used with a KL penalty coefficient of 0.05 and no entropy regularization. We adopt asymmetric PPO clipping with a clipping range of $[0.15,0.2]$.

Optimization uses Adam with $\beta_1=0.9$, $\beta_2=0.98$, and weight decay of 0.1 under a constant learning-rate schedule. Training runs for 200 rollout steps, with evaluation every 10 steps using greedy decoding.

Model-specific GRPO configurations are summarized in Table~\ref{tab:grpo-config}.

\begin{table}[t]
\centering
\caption{Model-specific GRPO training configurations.}
\label{tab:grpo-config}
\resizebox{0.5\textwidth}{!}{\begin{tabular}{lccc}
\toprule
\textbf{Setting}
& \textbf{Qwen3.5-9B}
& \textbf{Qwen3.5-27B}
& \textbf{Qwen3.5-35B-A3B} \\
\midrule
Tensor Parallel (TP) & 4 & 4 & 2 \\
Pipeline Parallel (PP) & 1 & 1 & 1 \\
Expert Parallel (EP) & 1 & 1 & 8 \\
Learning rate
& $1\times10^{-6}$
& $1\times10^{-6}$
& $5\times10^{-7}$ \\
Rollout batch size & 256 & 256 & 128 \\
Global batch size & 256 & 256 & 128 \\
Maximum response length & 4096 & 4096 & 3072 \\
Maximum tokens per GPU & 16384 & 16384 & 12288 \\
GPUs per rollout engine & 2 & 4 & 4 \\
SGLang static memory fraction & 0.50 & 0.60 & 0.45 \\
\bottomrule
\end{tabular}}
\end{table}

\section{Evaluation Details}

\subsection{Inference Settings}
\label{app:inference-settings}

For benchmark evaluation, all translation models use greedy decoding to ensure deterministic comparison. For hard-example mining and GRPO training, we use stochastic sampling with temperature 0.7 and top-$p$ 0.8.

\subsection{LLM-based Evaluation}
\label{app:llm-evaluation}

We use two LLM-based evaluators for cultural-effectiveness evaluation.

\paragraph{Agent-enhanced Judger.}

The Agent-enhanced Judger is the GRPO-aligned Judger described in Section~\ref{sec:agent-judger}. Given a source text, optional context, and candidate translation, it assigns a cultural-effectiveness score in $\{0,1,2,3\}$. When necessary, the retrieval Agent obtains external evidence and provides concise judging hints to the Judger. The Judger retains final scoring authority.

\paragraph{Gemini API Judger.}

Gemini uses the same cultural-effectiveness scoring rubric and evaluation prompt as CULTURE-MT~\citep{CULTURE-MT}. It is used as an external automatic evaluator in translation experiments.

Gemini also contributes to translation-data construction and Judger annotation during training. Therefore, its evaluation results should be interpreted as a complementary automatic signal rather than a fully independent evaluation. Human evaluation is used as the primary independent assessment.

For all automatic evaluators, cultural effectiveness is reported as the percentage of translations receiving scores of 2 or 3 under the four-level rubric.

\subsection{Human Evaluation}
\label{app:human-evaluation}

Human evaluation follows the four-level cultural-effectiveness rubric introduced in CULTURE-MT~\citep{CULTURE-MT}. Scores are defined as follows: 0 indicates severe loss of meaning or cultural intent; 1 indicates partial comprehensibility without effective cultural communication; 2 indicates generally effective translation with residual stylistic or cultural weaknesses; and 3 indicates highly effective translation.

Nine Chinese--English bilingual experts participate in evaluation. For each translation model, six experts are randomly assigned into two groups of three annotators. Within each group, annotators evaluate disjoint subsets of the 980 CUE-MT-Bench instances, ensuring that each instance receives one annotation from each group. Annotators are unaware of other group members and the evaluated model identity.

The two group-level annotations are first converted into binary labels:
\[
b(s)=\mathbb{I}[s\geq2].
\]
The binary agreement between the two groups is used to measure annotation consistency. The remaining three experts review disagreements and determine the final score in $\{0,1,2,3\}$.

The reported human cultural-effectiveness score is the percentage of final adjudicated scores equal to 2 or 3.

\subsection{Temporal Scope}
\label{app:temporal-scope}

The source texts used for Judger training and translation-model training were collected before November 30, 2025. The 980 instances in CUE-MT-Bench were sampled during April 30--May 30, 2026.

These dates specify the sampling periods of our task-specific datasets and do not represent the pretraining cutoff of the backbone models, whose training corpora and cutoff dates are not publicly available.

The benchmark contains both emerging and established social-media expressions. The motivation of this work is not limited to newly emerging cultural knowledge, but rather focuses on cases where required cultural knowledge is unavailable or insufficiently applied during evaluation.

\section{Judger Baselines and Ablations}
\label{app:judger-baselines}

\paragraph{Hint GRPO Baseline.}

We reimplement the Hint GRPO approach of \citet{qiu2026metricsrewardworsttranslations} on Qwen3.5-27B. Different from our Judger, which is initialized from the two-stage human-aligned SFT checkpoint, the Hint GRPO baseline starts directly from the base Qwen3.5-27B checkpoint. It uses the same 20K human-labeled difficult-example pool as our Judger GRPO training, and the judging hints are generated by GLM-5.2.

Therefore, this baseline evaluates hint-guided Judger optimization under a different initialization setting. The no-retrieval Agent ablation described below provides a matched comparison using the same GRPO-aligned Judger backbone while removing external evidence.

\paragraph{No-Retrieval Agent Ablation.}

The no-retrieval ablation in Table~\ref{tab:judger-main} retains the GRPO-aligned Judger, evaluation inputs, scoring pipeline, and the original Agent query-trigger decisions.

External retrieval is disabled. Only the inputs that originally trigger retrieval receive GLM-5.2-generated judging hints without retrieved evidence, while non-triggered inputs receive no hints. On CUE-Judge-Bench, 52.64\% of inputs trigger retrieval under the original Agent configuration.

This design preserves the same trigger population and isolates the contribution of retrieved external evidence from hint generation. Enabling retrieval improves binary accuracy and four-way accuracy by 3.86 and 8.07 percentage points, respectively, with gains of 0.0769 in $\kappa$ and 0.1035 in Spearman correlation.

\section{Qualitative Analysis of Cultural-Effectiveness Judging}
\label{app:judger-error-analysis}

We analyze three representative cases to understand how retrieval-enhanced judging affects contextual interpretation, error coverage, and severity calibration. Table~\ref{tab:judger-cases} reports human scores, Judger scores, and the corresponding cultural-effectiveness rewards used in translation training. These examples illustrate representative behaviors and limitations.

\begin{table}[htbp]
\centering
\caption{Selected human-annotated cases. Each cell reports score / implied $r_{\mathrm{culture}}$. Rewards are computed according to the translation-training mapping described in Section~\ref{sec:training-pipeline}. No claim is made that these examples were used for training.}
\label{tab:judger-cases}
\small
\resizebox{0.48\textwidth}{!}{\begin{tabular}{lccc}
\toprule
Example & Human & Static Judger & Agent-enhanced Judger \\
\midrule
843: Contextual slang & 3 / 1.0 & 2 / 1.0 & 3 / 1.0 \\
938: Named entity error & 2 / 1.0 & 3 / 1.0 & 1 / 0.1 \\
226: Abbreviation and intent error & 1 / 0.1 & 1 / 0.1 & 2 / 1.0 \\
\bottomrule
\end{tabular}}
\end{table}

\paragraph{Accepting a valid contextual rendering (Example 843).}

In a giveaway post, the source phrase ``\begin{CJK*}{UTF8}{gbsn}dd评论区和私信\end{CJK*}'' is translated as ``Drop a comment or DM me.'' The static Judger assigns 2, considering that the slang token ``dd'' is not explicitly preserved. The Agent-enhanced Judger interprets the phrase as a request for contact and assigns 3, consistent with the human score. In this context, the translation conveys the intended communicative function without explicitly translating the slang form. This example shows that contextual knowledge can prevent unnecessary penalties for acceptable adaptations.

\paragraph{Identifying an error with incorrect severity calibration (Example 938).}

The source title ``\begin{CJK*}{UTF8}{gbsn}无限暖暖修一下bug吧\end{CJK*}'' is translated as ``Fix the bug in Infinite Nikki please.'' Human annotators identify a game-title spelling error and assign 2. The static Judger overlooks the issue and assigns 3. The Agent-enhanced Judger retrieves the correct title, \emph{Infinity Nikki}, and identifies the entity error, but considers it overly severe and assigns 1. This example shows that improved error detection does not always guarantee better score calibration.

\paragraph{Correcting terminology while missing the overall intent (Example 226).}

The source ``\begin{CJK*}{UTF8}{gbsn}我买话剧票请看，去现场的话可以帮蹲一下sd吗\end{CJK*}'' is translated as ``I bought tickets to see the play; if anyone is going to the venue, could you please help me queue for the SD?'' Human annotators assign 1 because the translation changes the speaker's intent and leaves ``SD'' unresolved. The static Judger assigns 1 but incorrectly interprets SD as \emph{curtain call}. The Agent-enhanced Judger correctly identifies SD as \emph{Stage Door} and suggests expanding the abbreviation, but misses the broader intent error and assigns 2. This example highlights that local cultural knowledge retrieval improves terminology interpretation but does not always ensure complete sentence-level understanding.

Taken together, these cases show that cultural-effectiveness judging requires both contextual knowledge and accurate severity calibration. Retrieval can help recover missing cultural information and reduce some interpretation errors, while remaining challenges include comprehensive intent understanding and appropriate error weighting. These qualitative examples complement aggregate results without establishing that retrieval alone causes the observed improvements.

\section{Full FLORES-zh2en Results}
\label{app:flores-results}

{Table~\ref{tab:flores} provides the absolute scores underlying Figure~\ref{fig:flores}; each plotted change is the score difference from the corresponding Qwen3.5 base model.}

\begin{table}[htbp]
\centering
\caption{
{General-domain translation results on FLORES-zh2en. We report Agent and Gemini judgments together with reference-based BLEU, chrF, and COMET scores.}
}
\label{tab:flores}
\small
\resizebox{0.48\textwidth}{!}{\begin{tabular}{lccccc}
\toprule
\textbf{Model}
& \textbf{Agent}
& \textbf{Gemini}
& \textbf{BLEU}
& \textbf{chrF}
& \textbf{COMET} \\
\midrule

\rowcolor{gray!12}
\multicolumn{6}{l}{\textit{Model scale: 9B}} \\

Qwen3.5-9B
& 94.96 & 95.95 & 26.83 & 57.54 & 87.32 \\

CUE-MT w/o GRPO
& 93.67 & 93.37 & 22.96 & 54.11 & 85.82 \\

\textbf{CUE-MT}
& 94.66 & 93.48 & 21.09 & 52.55 & 85.22 \\

\midrule

\rowcolor{gray!12}
\multicolumn{6}{l}{\textit{Model scale: 27B}} \\

Qwen3.5-27B
& 97.04 & 98.52 & 28.72 & 59.06 & 87.75 \\

CUE-MT w/o GRPO
& 95.36 & 96.44 & 23.82 & 54.43 & 85.76 \\

\textbf{CUE-MT}
& 95.65 & 96.84 & 24.22 & 54.99 & 86.23 \\

\midrule

\rowcolor{gray!12}
\multicolumn{6}{l}{\textit{Model scale: 35B-A3B}} \\

Qwen3.5-35B-A3B
& 96.54 & 98.12 & 28.64 & 58.87 & 87.64 \\

CUE-MT w/o GRPO
& 93.53 & 92.89 & 22.92 & 53.67 & 85.05 \\

\textbf{CUE-MT}
& 94.07 & 94.86 & 22.40 & 53.62 & 85.96 \\

\bottomrule
\end{tabular}}
\end{table}

\section{Data Cleaning}
\label{app:data-cleaning}

We cleaned the collected data through three rounds of rule-based and LLM-assisted filtering. In each round, we first applied deterministic rules to anonymize explicit identifiers, including usernames and personal names, remove identifiable information where applicable, and filter offensive or harmful content. We then used GLM-5 to identify additional potentially problematic content that might not be captured by predefined rules. GLM-5 assisted content identification rather than making unrestricted filtering decisions. After each round, we manually inspected a sample of the cleaned data and added rules to address recurring issues identified during inspection. The updated rules were applied in subsequent cleaning rounds. After the third round, no harmful content was identified in the manually inspected sample, corresponding to an observed rate of 0\%.

\end{document}